\documentclass[runningheads]{llncs}
\usepackage[pdftex]{graphicx}
\usepackage{amsmath,amssymb}
\newcommand{\bm}[1]{\mbox{\boldmath $#1$}}
\usepackage{algpseudocode,algorithm}

\begin{document}

\title{A Trainable Multiplication Layer for Auto-correlation and Co-occurrence Extraction\thanks{Supported by Qdai-jump Research Program and JSPS KAKENHI Grant Number JP17K12752 and JP17K19402.}} 
\titlerunning{A Trainable Multiplication Layer} 


\author{Hideaki Hayashi\inst{1}\orcidID{0000-0002-4800-1761} \and
Seiichi Uchida\inst{1}\orcidID{0000-0001-8592-7566}}
%

\authorrunning{H. Hayashi and S. Uchida} 


\institute{Kyushu University, 744 Motooka Nishi-ku Fukuoka, JAPAN　\\
\email{\{hayashi, uchida\}@ait.kyushu-u.ac.jp}}

\maketitle

\begin{abstract}
In this paper, we propose a trainable multiplication layer (TML) for a neural network that can be used to calculate the multiplication between the input features. 
Taking an image as an input, the TML raises each pixel value to the power of a weight and then multiplies them, 
thereby extracting the higher-order local auto-correlation from the input image. 
The TML can also be used to extract co-occurrence from the feature map of a convolutional network.
The training of the TML is formulated based on backpropagation with constraints to the weights, 
enabling us to learn discriminative multiplication patterns in an end-to-end manner. 
In the experiments, the characteristics of the TML are investigated by visualizing learned kernels and the corresponding output features. 
The applicability of the TML for classification and neural network interpretation is also evaluated using public datasets.

\end{abstract}
{\allowdisplaybreaks
	\section{Introduction}
	Typified by the success of convolutional neural networks (CNNs) in recent computer vision research, 
studies using deep learning-based methods have demonstrated success in several fields 
\cite{ma2016learning}, \cite{greenspan2016guest}, \cite{zhang2016learning}, \cite{amodei2016deep}, \cite{majumder2017deep}.
Unlike the traditional machine learning techniques based on handcrafted features, 
these deep learning-based methods automatically extract features from raw input data via end-to-end network learning.

To exploit the end-to-end learning capability of the deep neural networks, 
numerous studies have investigated the development of a network layer that represents a certain function by incorporating a model into a layer structure \cite{wang2017g}, \cite{shih2017deep}, \cite{hayashi2015recurrent}. 
For example, Wang {\it et al.} \cite{wang2017g} proposed a trainable structural layer called a global Gaussian distribution embedding network, 
which involves global Gaussian \cite{nakayama2010global} as an image representation.

Despite such challenging efforts, 
the majority of layer structures are based on inner products of the input features and weight coefficients. 
There has been minimal research attempting to introduce multiplication of the input features. 
In classical pattern recognition techniques, however, 
multiplication of the input features is important because it represents auto-correlation or co-occurrence of the input features. 

This paper proposes a trainable multiplication layer (TML) for a neural network that can be used to calculate the multiplication between the input features. 
Taking an image as an input, the TML raises each pixel value to the power of a weight and then multiplies them, 
thereby extracting the higher-order local auto-correlation from the input image. 
The TML can also be used to extract co-occurrence from the feature map of a CNN.
The training of the TML is formulated based on backpropagation with constraints to the weights, 
enabling us to learn discriminative multiplication patterns in an end-to-end manner. 

The contributions of this work are as follows: 
\begin{itemize}
	\item A trainable multiplication layer is proposed.
	\item The learning algorithm of the TML based on constrained optimization is formulated.
	\item Applicability to classification and network interpretation is demonstrated via experiments.
\end{itemize}

	\section{Related Work}
	\subsection{Multiplication in a Neural Network Layer}
The majority of the existing neural network layers focus on the multiplication of the input features and weight coefficients; 
there are limited studies that discuss a network layer that calculates the multiplication of the input features. 
Classically, Sigma-Pi Unit based on the summation of multiplication is applied to a self-organizing map \cite{weber2007self}, \cite{valle2005elman}. 
Incorporation of probabilistic models such as a Gaussian mixture model into the neural network structure 
consequently leads to the development of a network layer based on multiplication
\cite{hayashi2015recurrent}, \cite{tsuji1999log}, \cite{tsuji2003recurrent}. 
In recent work, Shih {\it et al.} \cite{shih2017deep} proposed a network layer that detects co-occurrence by calculating multiplications between the feature maps of a CNN.
Kobayashi \cite{kobayashi2018trainable} achieved a trainable co-occurrence activation by decomposing and approximating a multiplication operation.

Although such multiplication-based layers are not actively studied, 
they have the potential to create an efficient network structure 
because they can express auto-correlation and co-occurrence directly. 
This paper therefore focuses on the development of the multiplication-based neural network layer.

\subsection{Higher-Order Local Auto-correlation}
Higher-order local auto-correlation (HLAC) is a feature extraction method proposed by Otsu \cite{otsu1988new}.
HLAC is frequently used in the field of image analysis \cite{uehara2017object}, \cite{fujino2014liver}, \cite{nosato2014objective}, \cite{hu2013modified}. 
For $L$ displacements $\bm{a}_1, \ldots, \bm{a}_L$, 
the $L$th-order autocorrelation function is defined as
\begin{eqnarray}
R_L(\bm{a}_1, \ldots, \bm{a}_L) &=& \sum_{D'} f(\bm{r})f(\bm{r}+\bm{a}_1) \cdots f(\bm{r}+\bm{a}_L), \\
D' &=& \{\bm{r} | \bm{r}+\bm{a}_l \in D, \forall l \in \{1, 2,\ldots, L \} \},
\end{eqnarray}
where $f(\bm{r})$ is the pixel value of the input image at a coordinate $\bm{r}$, 
$D$ is a set of coordinates of the input image. 

In HLAC, the patterns of the displacement must be prepared manually. 
The number of displacement patterns increases explosively based on the mask size and order; 
hence, they are limited practically.

	\section{Trainable Multiplication Layer}
	\subsection{Layer Structure}
Fig. \ref{fig:LayerStructure} presents an overview of the forward calculation in the TML. 
\begin{figure*}[t]
	\centering
	\includegraphics[width=1.0\hsize] {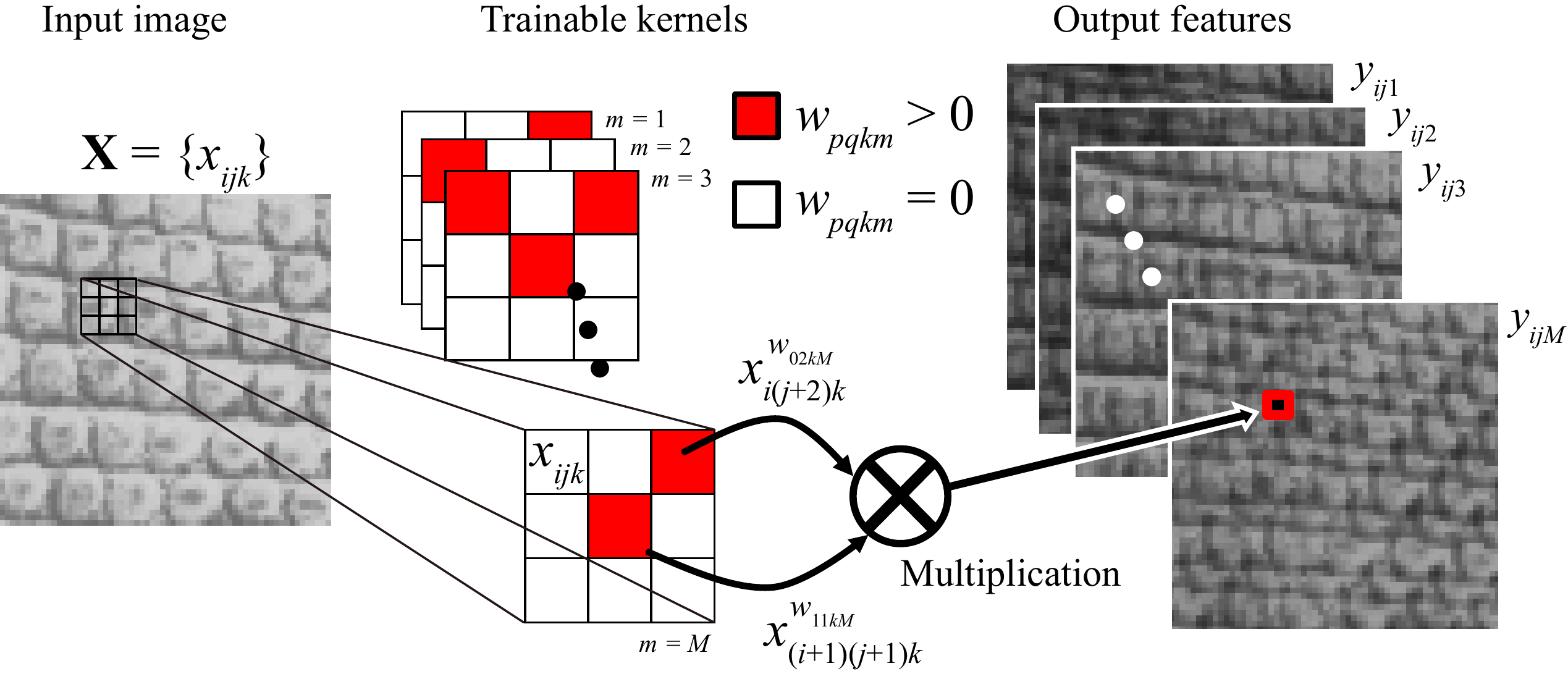}
	\caption{Overview of forward calculation conducted in the TML. 
					This figure shows a simple case where the number of input channels $K$ is $1$ and the kernel size is $3 \times 3$.}
	\label{fig:LayerStructure}
\end{figure*}
The main idea behind this layer is to achieve the multiplication of the input values chosen by the kernels, 
which is different from the well-known convolutional layer that conducts multiplication of the input values and kernels. 

Given an input image or feature map of a convolutional layer ${\bf X} \in \mathbb{R}^{N_1 \times N_2 \times K}$, 
where $N_1$, $N_2$, and $K$ are the number of rows, columns, and channels, respectively, 
the forward pass of the TML is defined as follows: 
\begin{equation}
y_{ijm} = \prod^{K-1}_{k=0}\prod^{H-1}_{p=0}\prod^{W-1}_{q=0} x_{(i+p)(j+q)k}^{w_{pqkm}}, 
\label{eq:forward}
\end{equation}
where $x_{ijk}$ ($i = 0, \ldots, N_1-1$, $j = 0, \ldots, N_2-1$, $k = 0, \ldots, K-1$) is the ($i, j, k$)-th element of ${\bf X}$ 
and $w_{pqkm} \geq 0$ is the ($p, q$)-th weight of the $m$-th kernel for channel $k$ 
($p = 0, \ldots, H-1$, $q = 0, \ldots, W-1$, $m = 0, \ldots, M-1$; $M$ is the number of kernels). 
Since any number to the zero power is one
(we define the value of zero to the power of zero to also be one), 
the forward pass of the proposed layer is regarded as the multiplication of the input values, 
where the value of the kernel at the corresponding coordinate is greater than zero.

In practice, (\ref{eq:forward}) is calculated in the logarithmic form to prevent under/overflow. 
Assuming that $x_{ijk} > 0$, since ${\bf X}$ is an image or a feature map passed through a ReLU function, 
(\ref{eq:forward}) is rewritten as follows:
\begin{eqnarray}
z_{ijmpqk} &=& \log \left( x_{(i+p)(j+q)k} + \epsilon \right), \\
y_{ijm} &=& \exp \left(\sum^{K-1}_{k=0}\sum^{H-1}_{p=0}\sum^{W-1}_{q=0} w_{pqkm} z_{ijmpqk} \right), 
\end{eqnarray}
where $\epsilon$ is a small positive quantity to avoid $\log0$. 

\subsection{Usage of the Layer in a Convolutional Neural Network}
\label{Section:Usage}
The TML is used as a part of a CNN-based deep neural network. 
Fig. \ref{fig:LayerUsage} shows two different usages of the TML in a CNN. 
\begin{figure*}[t]
	\centering
	\includegraphics[width=1.0\hsize] {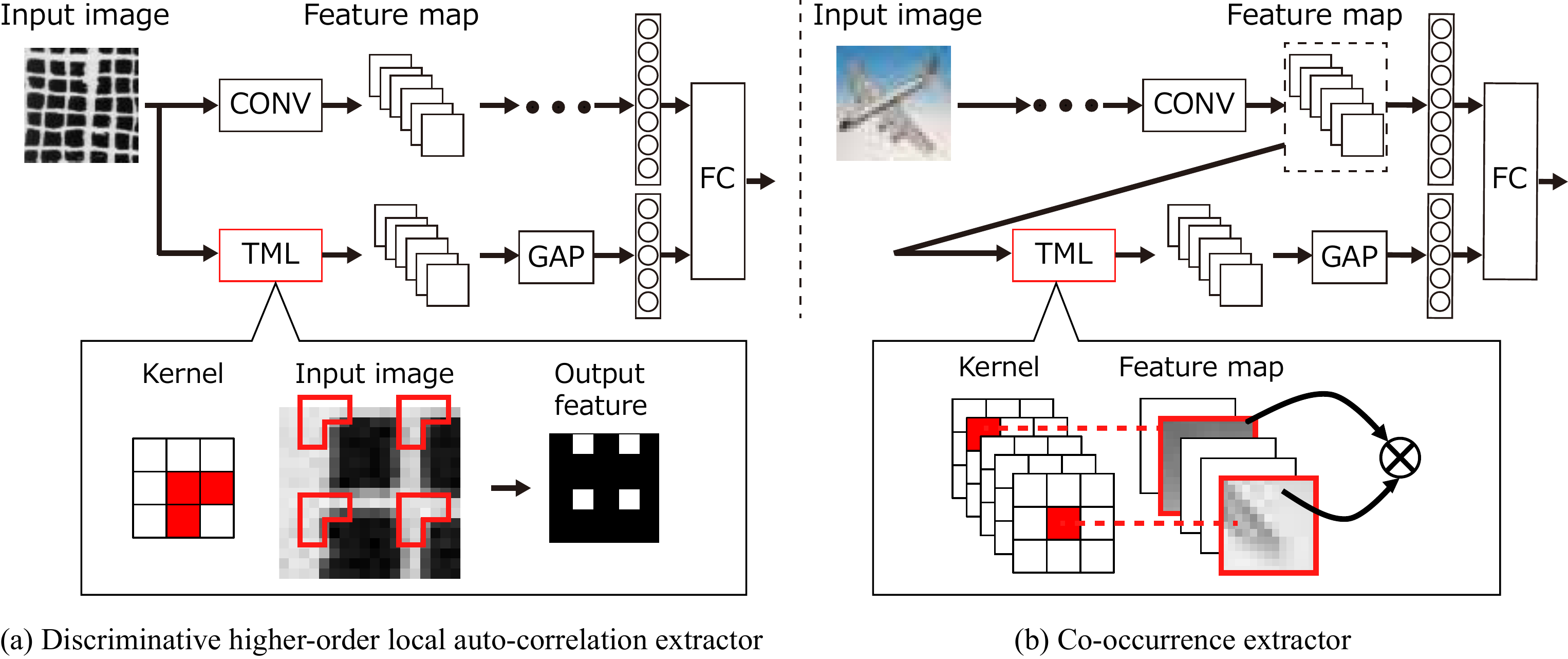}
	\caption{Two usages of the TML in a CNN: 
					(a) used as discriminative higher-order local auto-correlation extractor and 
					(b) used as co-occurrence extractor. 
			The abbreviations GAP, CONV, and FC denote 
			global average pooling, convolutional layer, 
			and fully connected layer, respectively. }
	\label{fig:LayerUsage}
\end{figure*}
Specifically, Fig. \ref{fig:LayerUsage}(a) shows the case that 
the TML is the use as a discriminative higher-order local auto-correlation (DHLAC) extractor; 
Fig. \ref{fig:LayerUsage}(b) shows the use as a co-occurrence extractor. 

The former usage is proposed to achieve the discriminative learning of the displacement patterns in HLAC. 
In this usage, the TML is inserted immediately behind the input image. 
The output features calculated by the TML are then passed through a global average pooling (GAP) \cite{Lin2013Network}.
GAP calculates the average of each feature map and the resulting vector is input to the next layer. 
This GAP is placed to simulate the integral computation of HLAC. 
Because the displacement patterns are represented by the power functions to be differentiable, 
they are trainable as kernel patterns.

The latter usage is designed to calculate the co-occurrence between feature maps. 
In the deeper layer of a CNN, feature maps calculated by a convolutional layer involve abstracted information of the input. 
In this usage, the TML determines the co-occurrence from this abstracted information 
by calculating the multiplication between the feature maps. 
This calculation allows the extraction of non-Markovian co-occurrence that is useful for classification. 
Furthermore, we can use the TML to extract not only co-occurrence between features of a single network, but also co-occurrence between two or more networks.

\subsection{Learning Algorithm}
Given a set of images ${\bf X}^{(n)}$ ($n = 1, \ldots, N$) for training with the teacher vector $\bm{t}^{(n)}$, 
the training process of the network into which the TML is incorporated 
involves minimizing the energy function $E$ defined as
\begin{align}
	\text{minimize } &\null E = \frac{1}{N}\sum^N_{n=1}{\rm Loss}(\bm{O}^{(n)}, \bm{t}^{(n)}) + \lambda \|\bm{w}\|_1 \label{eq:Energy} \\
	\text{subject to } &\null \sum^{K-1}_{k=0}\sum^{H-1}_{p=0}\sum^{W-1}_{q=0}w_{pqkm} = C_1, \label{eq:SumNorm} \\ 
					   &\null 0 \leq w_{pqkm} \leq C_2, \label{eq:WeightClip} 
\end{align}
where ${\rm Loss}(\bm{O}^{(n)}, \bm{t}^{(n)})$ is the loss function 
defined by the final network output vector $\bm{O}^{(n)}$ corresponding to ${\bf X}^{(n)}$ and the teacher vector $\bm{t}^{(n)}$. 
The operator $\| \cdot \|_1$ indicates the $L_1$ norm, $\bm{w}$ is all of the kernels presented in a vectorized form, and $\lambda$ is a constant. 
The constants $C_1$ and $C_2$ are hyperparameters that influence the training results, 
and the details are described in the next section. 
The $L_1$ regularization is recruited expecting to obtain sparse kernels based on the concept of the TML. 
The first constraint shown in (\ref{eq:SumNorm}) sets the total value of each kernel and prevents overflow. 
The second constraint shown in (\ref{eq:WeightClip}) prevents each kernel from being overly sparse because each kernel must have two or more nonzero elements to be a meaningful kernel. 
It could appear that the $L_1$ regularization in (\ref{eq:Energy}) is redundant as the $L_1$ norm is fixed to a constant value by both constraints. 
However, the $L_1$ regularization influences the gradient vectors to obtain a sparse solution during the network learning. 

Algorithm \ref{al:WeightUpdating} shows the procedure to solve the above optimization, 
where $\bm{w}_m$ is the vectorized $m$-th kernel. 
\begin{algorithm}[t]
	\caption{Algorithm of weight updating}
	\label{al:WeightUpdating}
	\begin{algorithmic}[1]
	\Require{Parameters $C_1$, $C_2$, and $\lambda$, training image set ${\bf X}^{(n)}$, teacher vector $\bm{t}^{(n)}$.}
	\Ensure{Trained network.}
	\State{Initialize the kernels $\bm{w}$ and other network weights $\bm{\theta}$.}
	\While{\bm{w} and \bm{\theta} have not converged}
		\State{Calculate $E$.}
		\State{Calculate gradients of $E$ with respect to $\bm{w}$ and $\bm{\theta}$.}
		\State{Update $\bm{w}$ and $\bm{\theta}$ using gradient-based updating.}
		\State{$\bm{w} \gets$ clip($\bm{w}, 0, C_2$)}
		\State{$\bm{w}_m \gets C_1\bm{w}_m/{\rm sum}(\bm{w}_m)$ for $m = 0, \ldots, M-1$}
	\EndWhile
	\end{algorithmic}
\end{algorithm}
In this algorithm, gradient-based weight updating to decrease the energy function $E$ 
and weight modification to maintain the constraints are calculated alternately. 
Although this is an approximated approach different from the well-known Lagrange multiplier, 
we employ this approach to strictly satisfy the constraints during the network learning.

To calculate backpropagation, the partial derivative of $E$ with respect to each kernel $w_{pqkm}$ is required. 
Since the TML consists of multiplication and power functions, 
the partial derivative is calculated simply as follows: 
\begin{eqnarray}
\frac{\partial E}{\partial w_{pqkm}}
&=& \frac{\partial E}{\partial y_{ijm}} \frac{\partial y_{ijm}}{\partial w_{pqkm}} \nonumber \\
&=& \frac{\partial E}{\partial y_{ijm}} \prod^{K-1}_{k=0}\prod^{H-1}_{p=0}\prod^{W-1}_{q=0} x_{(i+p)(j+q)k}^{w_{pqkm}} \log{x_{(i+p)(j+q)k}} \nonumber \\
&=& \frac{\partial E}{\partial y_{ijm}} y_{ijm} \log{x_{(i+p)(j+q)k}}, 
\end{eqnarray}
where the form of $\frac{\partial E}{\partial y_{ijm}}$ depends on the layer connected between the TML and the output. 

As outlined above, the TML incorporated in the deep network calculates DHLAC of the input image or co-occurrence between the feature maps. 
The kernels are trained in an end-to-end manner via backpropagation with constraints.

	\section{Investigation of the Layer Characteristics}
	Before beginning the classification experiments, 
we investigated the characteristics of the TML 
by training a relatively shallow network with the TML inserted, 
and visualizing the learned kernels and corresponding feature maps.

\subsection{Relationships between Hyperparameters and Learned Kernels}
The TML has important hyperparameters including $C_1$, $C_2$ for the learning constraints, and the kernel size. 
We investigated the changes in the learned kernels and the corresponding features according to the parameter variation.

In this trial, LeNet-5 \cite{lecun1998gradient} was used as the basic structure. 
The TML was inserted immediately behind the input.
The ReLU activation function \cite{nair2010rectified} was used for the convolution layers and the sigmoid function was used for the fully connected layer. 

For the training data, we used the MNIST dataset \cite{lecun1998gradient}. 
This dataset includes ten classes of handwritten binary digit images with a size of $28 \times 28$;  
it contains 60,000 training images and 10,000 testing images. 
The network was trained with all the training data for each hyperparameter. 

Following the network training, we visualized the learned kernels and the responses for the testing images. 
First, we observed the changes in the learned kernels according to the parameters for the constraints. 
Because the ratio of $C_1$ to $C_2$ influences the training results, 
we varied the value of $C_2$ as $C_2 = 1.0, 0.5, 0.33, 0.25$ for fixed a $C_1$ of $1.0$. 
The kernel size and $\lambda$ were set as $3 \times 3$ and $\lambda = 0.01$, respectively. 

Fig. \ref{fig:LearnedWeights} shows the changes in the learned kernels according to $C_2$. 
\begin{figure}[t]
	\centering
	\includegraphics[width=1.0\hsize] {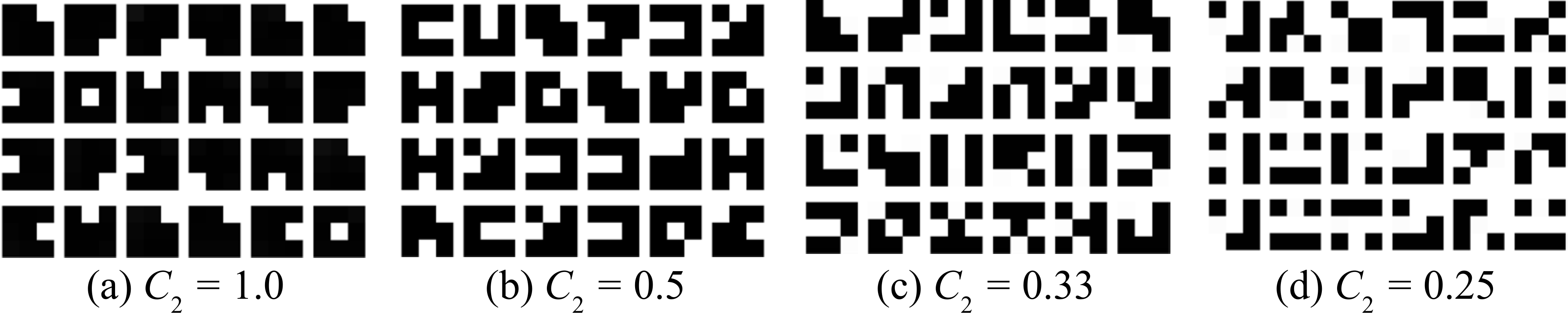}
	\caption{Changes in learned kernels according to $C_2$. 
					The kernel size and other parameters are fixed as $3 \times 3$, $C_1 = 1.0$, and $\lambda = 0.01$.}
	\label{fig:LearnedWeights}
\end{figure}
In this figure, the values of the learned kernels normalized by $C_2$ are displayed in a grayscale heat map, 
and therefore black and white pixels show ``0'' and $C_2$, respectively. 
The number of nonzero elements in each kernel increases according to the decrease in the value of $C_2$. 
Since the total value of elements in each kernel is fixed to $C_1$ and the upper limit of each element is suppressed by $C_2$, 
the number of nonzero elements approximates to $C_1/C_2$ if the $L_1$ regularization functions appropriately. 
These results demonstrate that the number of nonzero elements, which is equivalent to the order in HLAC, 
can be controlled by changing the ratio of $C_1$ to $C_2$.

Secondly, the kernel size was varied as $3 \times 3$, $5 \times 5$, $7 \times 7$, and $9 \times 9$ 
for the fixed value of $C_1 = 1.0$, $C_2 = 0.5$, and $\lambda = 0.01$.
Fig. \ref{fig:Responses} shows the learned kernels and the corresponding output features according to the kernel size. 
\begin{figure}[t]
	\centering
	\includegraphics[width=1.0\hsize] {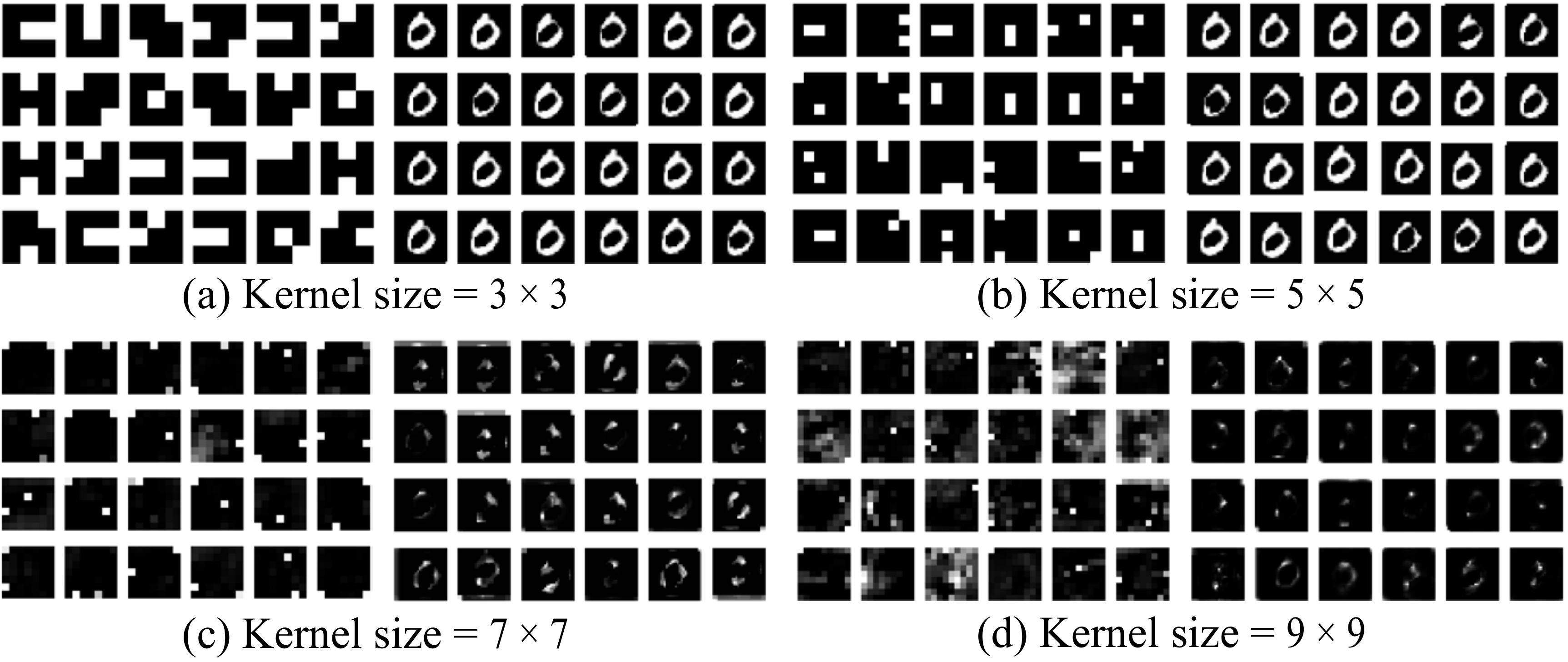}
	\caption{Learned kernels (left panels) and corresponding output features (right panels) for each kernel size. 
					Other parameters are fixed as $C_1 = 1.0$, $C_2 = 0.5$, and $\lambda = 0.01$. 
					}
	\label{fig:Responses}
\end{figure}
The kernel pattern and output feature pattern placed at the same location correspond to each other. 
The learned kernel values displayed in the left panels are also normalized in the same manner as Fig.  \ref{fig:LearnedWeights}. 
The values of the output features shown in the right panels are rescaled from [0, 1] to [0, 255] before visualization.

In terms of kernel size variation, the characteristics of the learned kernels and output features changed depending on the kernel size. 
In Fig. \ref{fig:Responses}(a), the number of nonzero elements is two in each kernel and these elements adjoin each other in the majority of the kernels. 
Kernels with neighboring nonzero elements extract rather local correlation, and the output features are virtually the same as the input images. 
However, kernels with nonzero elements apart from each other extract different output features according to the kernel pattern.
In Fig. \ref{fig:Responses}(b), the number of kernels with nonzero elements apart from each other increased and hence a richer variation of output features was obtained compared to Fig. \ref{fig:Responses}(a).
Fig. \ref{fig:Responses}(c) and (d) include kernels that have gray pixels. 
This means that three or more values are multiplied in a calculation and therefore high-order auto-correlation is extracted. 
These results indicate that the number of nonzero elements in each kernel can be controlled to a certain extent by the ratio of $C_1$ to $C_2$, 
and the variety of the output features changes according to the kernel size.

\subsection{Characteristics as a DHLAC Extractor}
We verified if the TML can make discriminative responses
by observing the learned kernel patterns and the corresponding output features for the synthetic texture patterns. 

The texture dataset used in this experiment contained six classes of $32 \times 32$ artificially generated images from the following procedure
(examples are displayed in the leftmost panels in Fig. \ref{fig:ToyProblemResults}): 
First, six stripe patterns, different for each class, were drawn on a $1024 \times 1024$ black image. 
Then, uniform random noise in the range of [0, 1] was added to the images. 
Finally, a set of randomly cropped $32 \times 32$-sized images were used as the dataset.
We generated 100 training images for each class (600 training images in total). 
After training, we observed the learned kernels and the corresponding output features 
to the testing samples that were generated independently from the training images.

The network structure used in this experiment was LeNet-5, as in Section 4.1. 
The parameters were set as $C_1 = 1.0$, $C_2 = 0.5$, $\lambda = 0.01$, and the kernel size was $9 \times 9$.

Fig. \ref{fig:ToyProblemResults} shows the output features for each combination of the input image and learned kernel.
\begin{figure}[t]
	\centering
	\includegraphics[width=0.71\hsize] {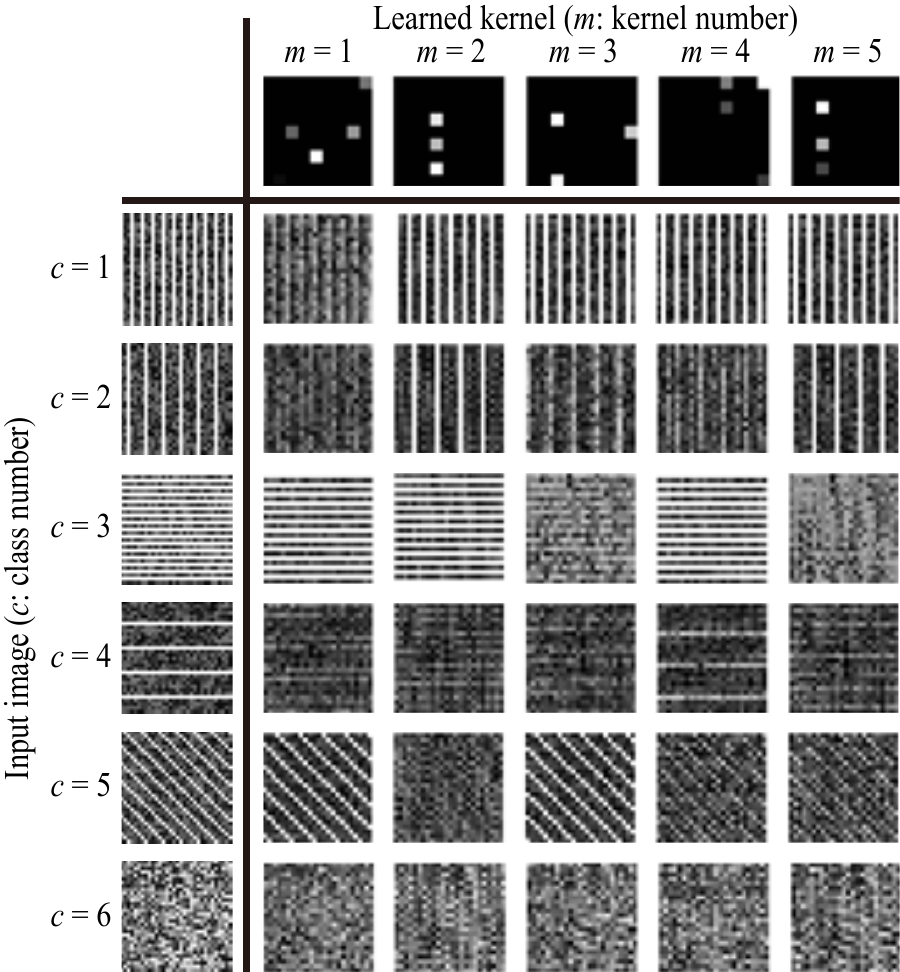}
	\caption{Output features for each combination of the input image and learned kernel.}
	\label{fig:ToyProblemResults}
\end{figure}
The row and column correspond to the class number and kernel number, respectively. 
As with Fig. \ref{fig:Responses}, the features are rescaled [0, 1] to [0, 255].

Fig. \ref{fig:ToyProblemResults} indicates that different output features are obtained according to the combination of the class and the kernel. 
For example, for the input image of $c = 5$, the slanting white stripe remains only in the output features from $m=1$ and $m=3$. 
This is because the kernels $m = 1$ and $m = 3$ have nonzero elements apart diagonally from each other. 
For the other classes, the pattern remains in the output features if the direction or interval of the pattern corresponds to those of the kernel patterns.
This means that the TML learned kernels that can extract discriminative features from the input image.

\subsection{Characteristics as a Co-occurrence Extractor}
To investigate the capability of the TML for extracting co-occurrence features, 
we visualized the co-occurring regions detected by the TML. 

For the use as a co-occurrence extractor, 
the TML was connected to the fully connected layer as shown in Fig. \ref{fig:LayerUsage}(b). 
In this experiment, the TML was inserted between the second convolutional layer and the fully connected layer of the LeNet-5.
Because this fully connected layer maps the input vector into the likelihood for each class, 
the weights of this layer represent the relevance between each dimension of the input vector and each class. 
Based on this fact, we extracted the most relevant kernel to the target category. 
We then defined the co-occurrence features as the input features to the TML 
that were activated by the kernel relevant to the target category.

Fig. \ref{fig:coocmap} shows the visualization results of the co-occurrence features.
\begin{figure}[t]
	\centering
	\includegraphics[width=1.0\hsize] {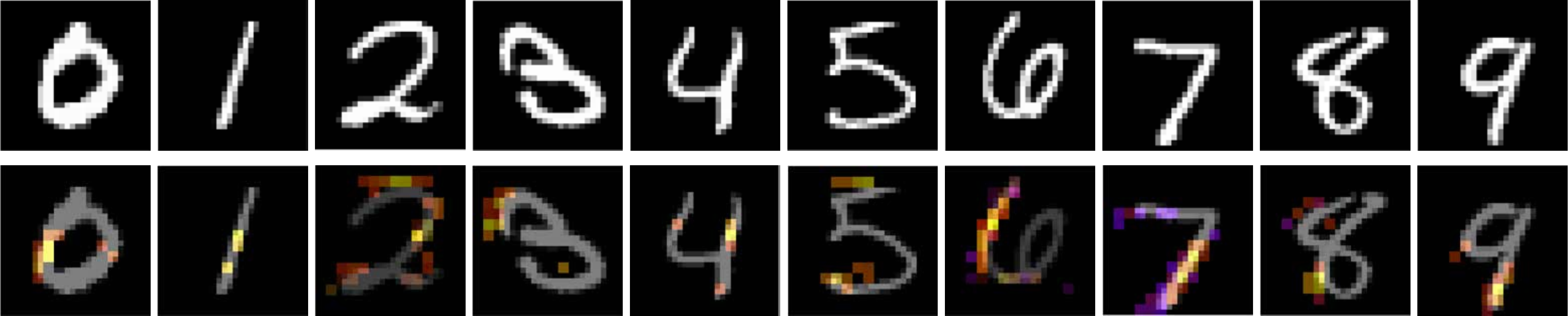}
	\caption{Visualization of the co-occurrence features. The top panels show examples of the original MNIST images. 
					The bottom panels are the corresponding co-occurrence features highlighted over the original images.}
	\label{fig:coocmap}
\end{figure}
In this figure, it appears that the distinctive features for each category are highlighted. 
For example, the bilateral lines of ``0'' are highlighted. In ``7'' the upper left edge and vertical line are highlighted.
These results demonstrate the capability of the TML for extracting co-occurrence features that are useful for classification from CNN feature maps.

	\section{Classification Experiments}
	To evaluate the applicability of the TML, 
we conducted a classification experiment using public datasets. 

\subsection{Dataset}
We used the following datasets in this experiment.

{\bf MNIST}: 
As described in the previous section, this dataset includes ten classes of handwritten binary digit images with a size of $28 \times 28$.  
We used 60,000 images as training data and 10,000 images as testing data. 

{\bf Fashion-MNIST}: 
Fashion-MNIST \cite{xiao2017} includes ten classes of binary fashion images with a size of $28 \times 28$.  
It includes 60,000 images for training data and 10,000 images for testing data. 

{\bf CIFAR10}: 
CIFAR10 \cite{krizhevsky2009learning} is labeled subsets of the 80 million tiny images dataset. 
This dataset consists of 60,000 $32 \times 32$ color images in 10 classes, with 6,000 images per class. 
There are 50,000 training images and 10,000 test images. 

{\bf Kylberg}: 
The Kylberg texture dataset \cite{Kylberg2011c} contains unique texture patches for each class. 
We used its small subset provided by the author which includes six classes with 40 samples each. 
We divided the original $576 \times 576$ patches into nine $64 \times 64$ nonoverlapping patches
and considered each patch as one sample; 
thus 2,160 samples were available. 
We conducted 10-fold cross-validation and calculated the classification accuracy. 

{\bf Brodatz}: 
The Brodatz texture dataset \cite{brodatz1966textures} 
contains 112 texture classes with a size of $640 \times 640$. 
We divided each image into 100 $64 \times 64$ nonoverlapping patches, 
and considered each patch as one sample; 
thus 11,200 samples were available. 
We conducted 10-fold cross-validation to calculate the accuracy. 

\subsection{Experimental Setup}
As the baseline, we used a simple CNN (called ``baseline CNN'' hereinafter) 
to clarify the effect of the TML. 
The baseline CNN consisted of five convolutional layers with four maxpooling layers between them; it also included two fully connected layers (the structure is illustrated in the supplemental file). 
In the network, dropout was conducted after the first and second maxpooling and the first fully connected layer.

The effectiveness of the TML was then examined by connecting the TML to the baseline CNN, 
both as a DHLAC extractor and a co-occurrence extractor. 
The kernel size of the TML was set as $5 \times 5$. 
We also compared the results to the results using HLAC instead of the TML. 


\subsection{Results}
Table \ref{table:BrodatzResults} shows the recognition rates for each dataset.
\begin{table}[t]
	\centering
	\caption{Comparison of recognition rates (\%).}
	\scalebox{0.9}[0.9]{
	\begin{tabular}{llllll}
 											& MNIST 		& Fashion & CIFAR10		& Kylberg 	& Brodatz	\\ \hline \hline
Baseline CNN 								& 99.27  		& 92.44 		& 81.43 		& 99.12 		& 91.69	\\
Baseline CNN + HLAC					 		& 99.31 		& 92.09 		& {\bf 82.23} 	& 99.31 		& 91.33	\\
{\bf Baseline CNN + TML (DHLAC)} 			& {\bf 99.39}	& 92.45 		& 81.50 		& 99.02 		& {\bf 92.51}	\\
{\bf Baseline CNN + TML (Co-occurrence)}	& 99.27 		& {\bf 92.54} 	& 81.49 		& {\bf 99.35} 		& 91.56	\\ \hline
\end{tabular}
	}
	\label{table:BrodatzResults}
\end{table}
In all datasets, applying the TML to the CNN improved the classification performance. 
In particular, the use as DHLAC demonstrated remarkable performance for the Brodatz dataset. 
This is because texture is based on a repeated structure, and hence auto-correlation information is effective for the classification.  
Moreover, the discriminative training of the HLAC features conducted by the TML functioned efficiently. 
These results confirm the applicability of the TML for classification.

	\section{Applications}
	\subsection{Interpretation of the Network}
In this section, we demonstrate that the TML can be used to interpret 
on which part of an input image the CNN focuses. 
In the usage as a co-occurrence extractor, the TML is connected between 
the feature maps and the fully connected layer (Fig. \ref{fig:LayerUsage}(b)). 
By tracing this structure in the reverse direction in the trained network, 
it is possible to extract features having a strong influence for classification. 
It should be emphasized that the purpose of this experiment is not to improve classification accuracy, 
rather, it is to improve interpretability. 

Specifically, we interpret a trained network in the following procedure: 
1. Perform a forward calculation for a certain input image; 
2. Calculate the largest weight among the weights of the fully connected layer
between the unit that outputs the posterior probability of the target class 
and the features calculated by the TML; 
3. Extract the kernel of the TML connecting the largest weight calculated in the previous step; 
4. Extract CNN feature maps connecting to nonzero elements of the kernel; and 
5. Visualize the extracted feature maps by upsampling them to the size of the input image and overlapping them on the input image. 
The property of the TML that the learned kernel acquires sparsity 
allows this calculation.

We used the Caltech-UCSD Birds-200-2011 dataset (CUB-200-2011) \cite{WahCUB_200_2011}. 
Each species of birds has unique characteristics such as feather pattern and body shape. 
This fact makes the interpretation of visualization results easier.
This dataset contains 200 species of bird images 
and consists of 5,994 images for training data and 5,794 images for test data. 
Although this dataset is frequently used for fine-grained classification and segmentation, 
we used this dataset for only visualization in this experiment. 

The basic network structure used in this experiment was the LeNet-5. 
The TML was inserted between the second convolutional layer and the fully connected layer. 
The parameters were set as $C_1 = 1.0$, $C_2 = 0.5$, $\lambda = 0.01$, and $1 \times 1$ for the kernel size.

Fig. \ref{fig:Cub200} shows examples of the visualization results. 
\begin{figure}[t]
	\centering
	\includegraphics[width=1.0\hsize] {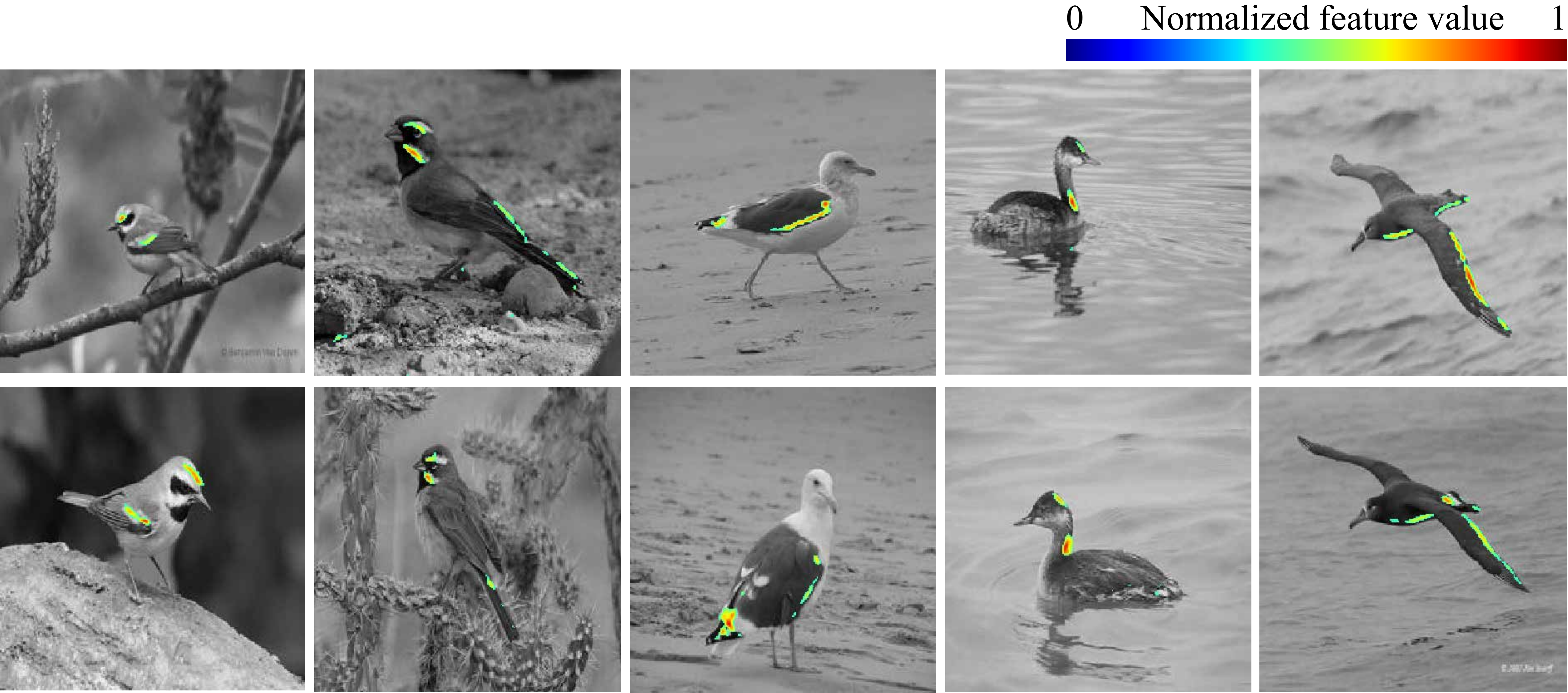}
	\caption{Visualization of network interpretation results using the TML. 
			The areas having a strong influence for classification extracted by the TML are highlighted.}
	\label{fig:Cub200}
\end{figure}
In Fig. \ref{fig:Cub200}, images of the same species are arranged in each column. 
The feather patterns or body shapes unique for each species are highlighted.  
For example, in the far left panels, yellowish green patterns on the head and the wing are highlighted. 
These results indicate the applicability of the proposed TML for interpreting a neural network. 

\subsection{Co-occurrence Extraction between Two Networks for Multimodal Data}
\label{Section:multimodal}
In this experiment, we applied the TML to multimodal data classification. 
As mentioned in Section \ref{Section:Usage}, the TML can also be used to extract co-occurrence between two networks. 
By extracting the co-occurrence between the features of two CNNs that take input data from different modalities, it is expected that the two networks complement each other. 
\begin{figure}[t]
	\centering
	\includegraphics[width=1.0\hsize] {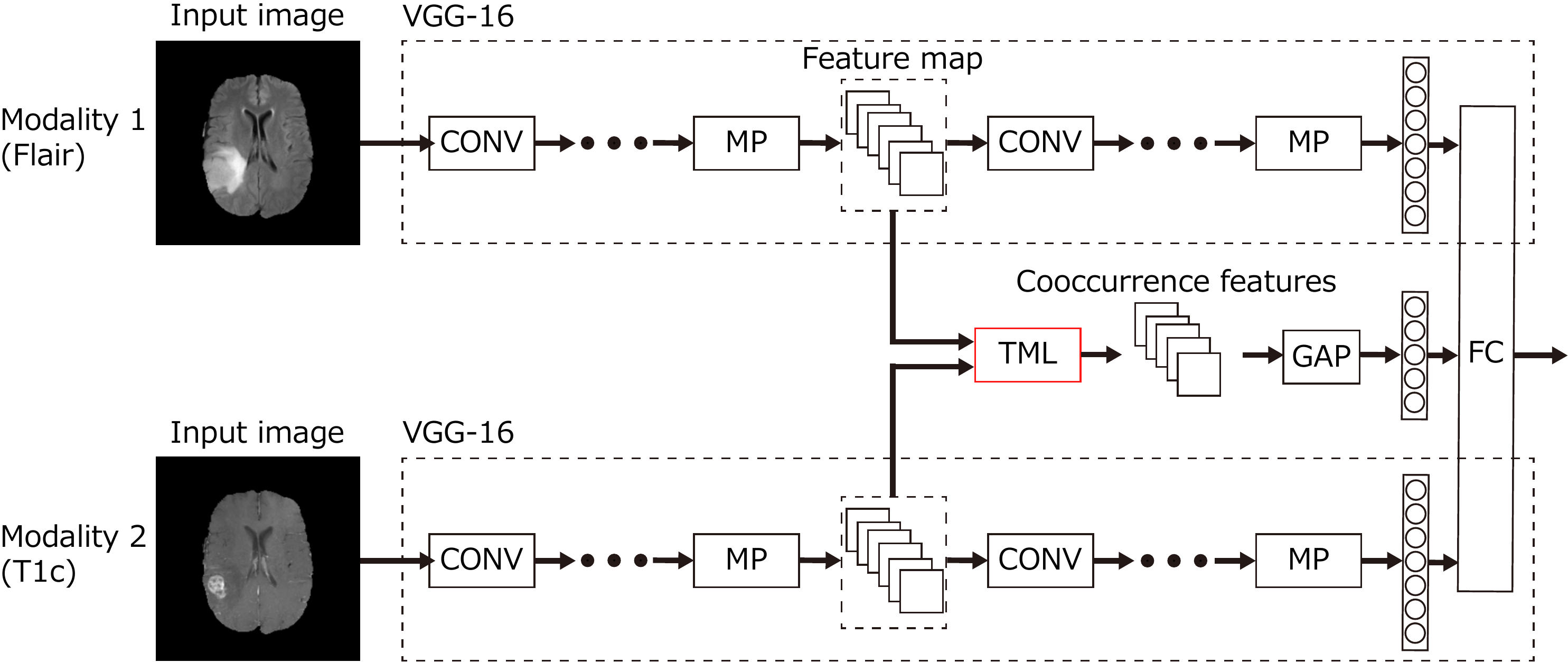}
	\caption{Network architecture for MRI classification. 
		The abbreviations MP, GAP, CONV, and FC denote 
		max pooling, global average pooling, convolutional layer, 
		and fully connected layer, respectively. 
		Each VGG-16 model takes different modality images as input.
		The co-occurrence features between two networks are extracted by the TML.}
	\label{fig:MRI_method}
\end{figure}

The problem we addressed in this experiment is tumor detection in magnetic resonance (MR) images. 
In MR imaging, there are different modalities depending on a setting of pulse sequences. 
Because each modality has a different response, 
extracting co-occurrence between multiple modalities could possibly improve tumor detection accuracy. 

We prepared an MR image dataset containing two modalities (Flair and T1c) from 
the multimodal brain tumor image segmentation benchmark (BraTS) \cite{menze2015multimodal}. 
This dataset consisted of three-dimensional brain MR images with tumor annotation. 
We created two-dimensional axial (transverse) slice images and separated them into tumor and non-tumor classes based on the annotation. 
The dataset contained 220 subjects; 
we randomly divided these into 154, 22, and 44 for training, validation, and testing, respectively. 
Because approximately 60 images were extracted from each subject, 
we obtained 8,980, 1,448, and 2,458 images for training, validation, and testing, respectively.
We resized the images from $240 \times 240$ pixels to $224 \times 224$ to fit the network input size. 

Fig. \ref{fig:MRI_method} illustrates the network architecture used in this experiment. 
The network was constructed based on VGG-16 \cite{simonyan2015very}. 
VGG-16 has three fully connected layers 
following five blocks consisting of convolutional layers and a max pooling layer. 
We applied the TML to the CNN features after the $b$-th max pooling layer from the top. 
For comparison, we calculated the classification accuracy for a single VGG-16 with each modality 
and two VGG-16's concatenated at the first fully connected layer. 

Table \ref{table:MRIResults} shows the results of the MR image classification. 
This confirms that the TML is effective for improving classification performance. 
In particular, the improvement is greatest for $b=3$. 
One possible explanation for this is that the information necessary for classification 
is extracted to the extent that position information is not lost in the middle of the network, 
and extracting co-occurrence from such information is effective for multimodal data classification. 
This result demonstrates the effectiveness of co-occurrence extraction using the TML.

\begin{table}[t]
	\centering
	\caption{Comparison of classification accuracies for the MRI dataset.}
	\begin{tabular}{llll}
		Method					& Modality			& Accuracy (\%)\\ \hline \hline
		Single VGG				& Flair 			& 94.22 \\ 
		Single VGG				& T1c		 		& 89.86 \\
		Concatenated two VGGs\hspace{5mm} & Flair and T1c\hspace{5mm} 	& 94.47 \\ 
		Ours ($b=1$)	 		& Flair and T1c 	& 95.16 \\
		Ours ($b=2$) 			& Flair and T1c 	& 95.57 \\
		Ours ($b=3$) 			& Flair and T1c 	& {\bf 96.14} \\
		Ours ($b=4$) 			& Flair and T1c 	& 95.85 \\
		Ours ($b=5$) 			& Flair and T1c 	& 95.77 \\ \hline
	\end{tabular}
	\label{table:MRIResults}
\end{table}

	\section{Conclusion}
	In this paper, we proposed a trainable multiplication layer (TML) for a neural network that can be used to calculate the multiplication between the input features. 
Taking an image as an input, the TML raises each pixel value to the power of a weight and then multiplies them, 
thereby extracting the higher-order local auto-correlation from the input image. 
The TML can also be used to extract co-occurrence from the feature map of a convolutional network.
The training of the TML is formulated based on backpropagation with constraints to the weights, 
enabling us to learn discriminative multiplication patterns in an end-to-end manner. 
In the experiments, the characteristics of the TML were investigated by visualizing learned kernels and the corresponding output features. 
The applicability of the TML for classification was also evaluated using public datasets. 
Applications such as network interpretation and co-occurrence extraction between two neural networks were also demonstrated.

In future work, 
we plan to investigate the practical applications of the proposed TML.
We will also expand the layer to the 3D structure in the same manner as cubic HLAC \cite{kobayashi2004action}. 
Application to time-series data analysis is also expected by constructing a one-dimensional structure.

}


\clearpage
\section*{Supplementary material}
\begin{figure}[h]
	\centering
	\includegraphics[width=\hsize] {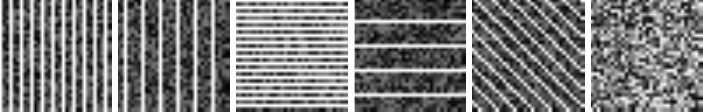}
	\caption{Examples of texture images used for the layer response observation}
	\label{fig:ToyProblemInput}
\end{figure}
The patterns of each class are as follows:
\begin{description}
\item Class 1: Vertical white stripe with two pixels of intervals 
\item Class 2: Vertical white stripe with four pixels of intervals
\item Class 3: Horizontal white stripe with one pixel of intervals
\item Class 4: Horizontal white stripe with six pixels of intervals
\item Class 5: Slanting white stripe with two pixels of intervals
\item Class 6: Nothing
\end{description}

\begin{figure}[h]
	\centering
	\includegraphics[width=1.0\hsize] {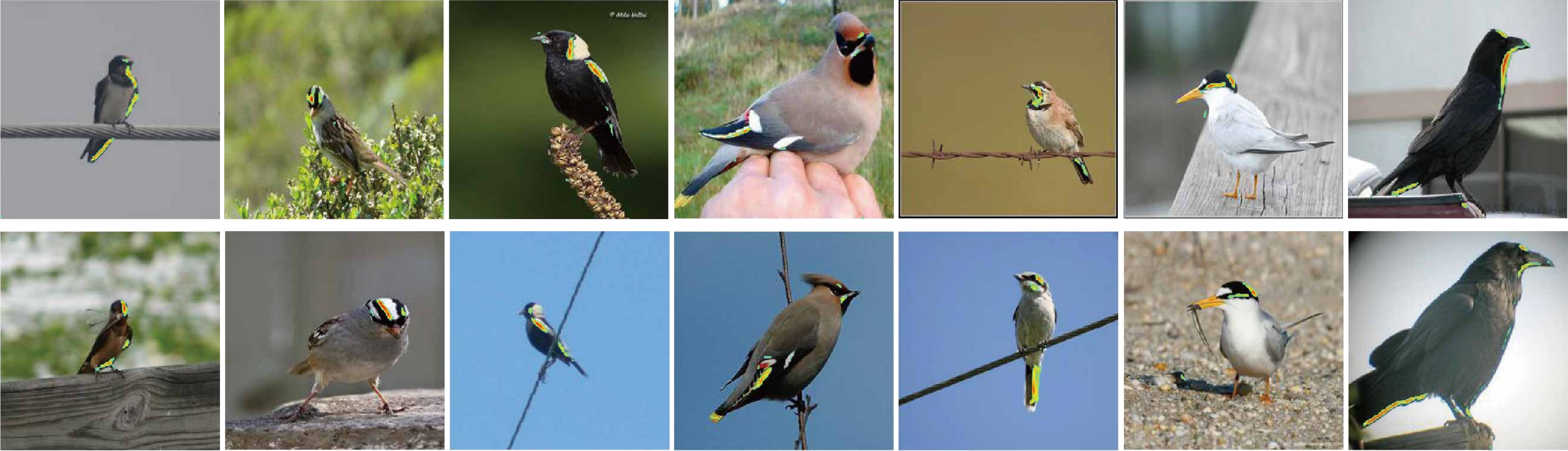}
	\caption{Other results of the network interpretation using the TML.}
	\label{fig:Cub200}
\end{figure}

\begin{figure*}[h]
	\centering
	\includegraphics[width=1.0\hsize] {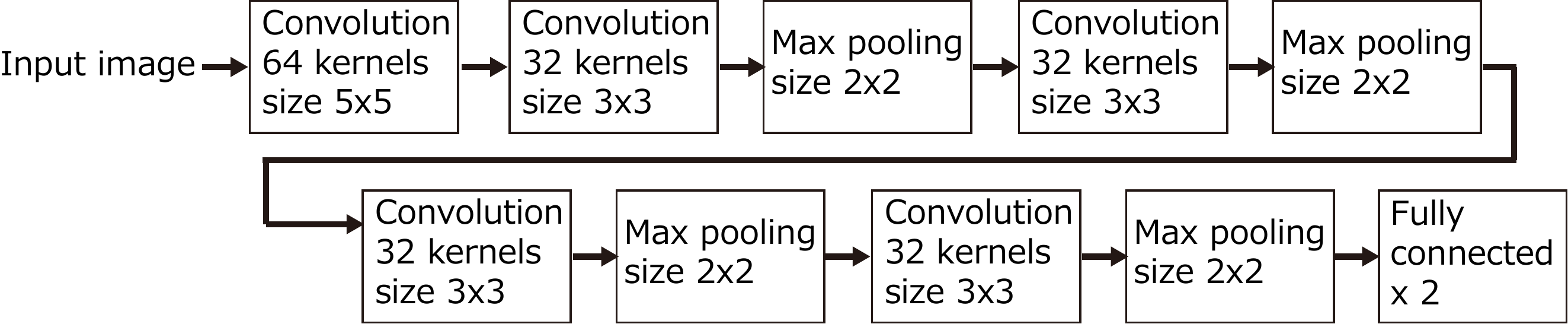}
	\caption{Structure of the baseline CNN. }
	\label{fig:netForBrodatz}
\end{figure*}

\end{document}